\documentclass[preprint,12pt]{elsarticle}

\usepackage{amssymb, amsthm,subfigure,amsmath,color}
\usepackage{algpseudocode}
\usepackage{algorithm}
\journal{Energy and Building}

\begin{document}

\begin{frontmatter}


\author{Rajib Rana\fnref{label1}}
\ead{to.rajib.rana@gmail.com}
\author{Brano Kusy\fnref{label1}}
\ead{brano.kusy@csiro.au}
\author{Josh Wall\fnref{label2}}
\ead{josh.wall@csiro.au}
\author{Wen Hu\fnref{label1}}
\ead{Wen.Hu@csiro.au}
%\fntext[label1]{{Autonomous Systems Program, CSIRO}}
%\cortext[cor1]{}
\address[label1]{Autonomous Systems Program, CSIRO}
\address[label2]{Intelligent Efficiency, CSIRO}
%\fntext[label3]{}

\title{Novel Methods for Activity Classification and Occupany Prediction Enabling Fine-grained HVAC Control.}


%\author{}
%
%\address{}

\begin{abstract}
Much of the energy consumption in buildings is due to HVAC systems, which has motivated several recent studies on making these systems more energy-efficient. Occupancy and activity are two important aspects, which need to be correctly estimated for optimal HVAC control. 
%These can maximize energy savings subject to preserving thermal comfort. 
However, state-of-the-art methods to estimate occupancy and classify activity require infrastructure and/or wearable sensors which suffers from lower acceptability due to higher cost. Encouragingly, with the advancement of the smartphones, these are becoming more achievable. Most of the existing occupancy estimation techniques have the underlying assumption that the phone is always carried by its user. However, phones are often left at desk while attending meeting or other events, which generates estimation error for the existing phone based occupancy algorithms. Similarly, in the recent days the emerging theory of Sparse Random Classifier (SRC) has been applied for activity classification on smartphone, however, there are rooms to improve the on-phone processing. We propose a novel sensor fusion method which offers almost 100\% accuracy for occupancy estimation. We also propose an activity classification algorithm, which offers similar accuracy as of the state-of-the-art SRC algorithms while offering 50\% reduction in processing. 
\end{abstract}


\end{frontmatter}


\section{Introduction}
\label{sec:intro}
HVAC is a dominant power consumer in commercial buildings. Much efforts have been invested in the past to efficiently control HVAC. Conventionally, most HVAC systems intake temperature and humidity to control cooling~\cite{ashrae2004standard}. This limitation can often lead to inefficient energy usage. For example, a room might be cooled to a conservative $24^oC$ even when it is unoccupied. Similarly, conventional HVAC control strategy does not take into consideration human comfort factors, such as recent physical activity. This can cause discomfort. For example, upon returning to desk after climbing many stairs, worker would like the temperature to be cooler than the conservative set point of $24^oC$. 

%
%regardless of whether there are any occupants.Out of many critical parameters, occupancy~\cite{Agarwal:2010:OEM:1878431.1878433} and human activity level~\cite{orosa2011new} have been identified as a crucial attributes that can help efficient and fine-grained HVAC control. 
A number of studies have been conducted in the past that attempted to 
estimate occupancy using sensors spanning motion sensors, RFID and contact sensors. However, introducing additional sensors incur cost of installation and maintenance, especially for large commercial buildings. To alleviate this problem, a number of studies have attempted to use smartphone for occupancy estimation. For example, \cite{krioukov2012personal,Balaji:2013:SOB:2517351.2517370,christensen2014using} uses WiFi signal strength to localize the phone. The general control strategy of these methods are, if the phone location is estimated as the office, the HVAC is activated, otherwise the local unit is turned off. However, note that if the phone is left on desk while the person is away, all of these systems will still determine the office space as occupied. This will lead to significant energy wastage. We overcome this challenge in our paper. 

A large body of literature can be found which use wearable sensors for activity detection. Given the current advancement in smartphone sensor technology, our objective is to piggyback the activity estimation on the smartphone to save additional expenses of wearable devices. In particular, we propose a \emph{Sparse Random Classifier (SRC)}~\cite{wright2009robust} based activity classifier, which pave the pathway to featureless classification that is suited to embedded smartphone platform. Notably, a number of attempts~\cite{zhang2013human,xu2012robust,liu2009human,xu2012co} have been taken in the past to use SRC for activity classification. However, the key limitations of these approaches is that higher accuracies are obtained when the feature dimension is significantly higher. Higher feature dimension warrants higher processing on embedded platforms, which could potentially interrupt the processing of other applications on the phone. In this paper we seek to address this challenge. We extend the theory of SRC and show that we can achieve similar accuracy as of the existing proposals, however, with half feature dimension.

Our contributions in this paper are as follows: 
\begin{enumerate}
\item We propose a sensor fusion method that uses sensor feed from phone microphone and phone accelerometer to determine office occupancy  when the phone is placed on the office desk. Privacy is a critical issue when using audio data. We preserve privacy by not storing or transiting raw audio data. Features are directly extracted from raw audio data and used for classification.

\item We also propose a extension of the Sparse Random Classifier for physical activity classification on smartphone. Our classifier achieves better accuracy with significantly smaller feature dimension. 

\item Experimental results show that our proposed fusion algorithm achieves 100\% accuracy in estimating office occupancy. Results also show that our activity classification algorithm offers greater than 95\% accuracy with 50\%
smaller feature dimension compared to the existing methods.
\end{enumerate}
%The state-of-the-art technique of measuring human comfort requires the measurement of sophisticated  parameters, such as clothing insulation,  efficient consumes a vast proportion of energy. Adaptive HVAC control is necessary to maximize energy savings. Two key factors to maximize energy saving include 1. \emph{occupancy estimation} and 2. \emph{activity inference}. Correctly measuring these two quantities require sophisticated hardware or software. For example, reed switch, etc was used to correctly measure the occupancy. On the other hand instantaneous metabolism rate (MET) can be determined from the activities. Previous studies have used user entered activity to determine MET. 
%
%
%Similarly, 
This paper is organized as follows. In the next section (Section~\ref{Sec:activity}), after providing some background on SRC, we describe our proposed extension of SRC for activity classification. Then in Section~\ref{sec:sensorFusion},  we first describe the Support Vector Regression model for occupancy inference using individual sensing modality. Then we describe our sensor fusion algorithm for occupancy estimation jointly using multi-modal sensing. We then present the experimental results in Section~\ref{sec:results} and finally conclude in Section~\ref{sec:conclude}.
% main text

%\section{Literature Review}
%We discuss the existing literature aligning with our contributions in this paper. In particular, we categorize the literature in two groups; first group discusses the  papers attempted to perform activity classification on smartphone. The second group encapsulates the papers discussing occupancy estimation using smartphone.
%
%\subsection{Activity Classificaiton Using Smartphone}
%
%\subsection{Occupancy Estimation Using Smartphone}
%In fact our system can be complementary to these existing systems by offering this additional degree of occupancy estimation. 

\section{Sparse Random Classifer for Activity Classifcation}
\label{Sec:activity}

\subsection{Sparse Random Classifier}
The Sparse Random Classifier(SRC) has been developed underpinning the theory of Compressive Sensing (CS)~\cite{wright2009robust}. SRC has been heavily used for classification in the past~\cite{sivapalan2011compressive,chew2012sparse,wei2013real,xu2011dynamic,Wei:2012:DSA:2185677.2185699}. The underlying assumption of SRC is that, given sufficient training samples of the $i$th activity class (such as walking or running and os on) $A_i = [v_{i,1} + v_{i,2}+ ... + v_{i,n_i}] \in \mathbb{R}^{m\times n_i}$, any test object $y\in \mathbb{R}^m$ from the same activity class will approximately lie in the linear span of the training samples associated with object $i$. Mathematical representation of this assumption using $\alpha_{i,j}$ as the coefficients can be given by \eqref{eqn:basis}.
\begin{eqnarray}
y = A_i = [\alpha_{i,1}v_{i,1} + \alpha_{i,2}v_{i,2}+ ... + \alpha_{i,n_i}v_{i,n_i}] \label{eqn:basis}
\end{eqnarray}
However, the membership $i$ of the test sample is unknown  primarily, we define a new matrix $A$ for the entire training set was the concentration of the all $n$ training samples of $k$ object classes:

\begin{eqnarray}
A = [A_1,A_2,...,A_k] = [v_{1,1} + v_{1,2}+ ... + v_{k,n_k}] 
\end{eqnarray}

Then the linear representation of the training object $y$ can be rewritten in terms of training samples as

\begin{eqnarray}
y = A x_0 \in \mathbb{R}^m
\end{eqnarray}
 
where $x_0 = [0,...,0,\alpha_{i,1} , ... , \alpha_{i,n_i},0,...,0] \in \mathbb{R}^n]$ is a coefficient vector whose entries are zero except those associated with $i$-class. 

The solution $x_0$ can be obtained by solving the system of equation $y = Ax$ when $m>n$. However, in reality $m<n$ (we will explain this later in this section), therefore the system is underdetermined. Conventionally, this difficulty is resolved by solving the minimum $\ell_2$-norm solution:
\begin{eqnarray}
\hat{x}_2 = \arg \min ||x||_2 \mbox{    s. t.   }   Ax = y, \label{eqn:l2norm}
\end{eqnarray}
 This optimization in~\eqref{eqn:l2norm} can be solved easily by pseudo inverse of A, however, the solution $\hat{x}$ is not informative. This is because we expect the solution to be sparse where only the coefficients related to test object class are non-zero. However, the $\ell_2$ norm provides a dense solution with many non-zero entries spanning multiple classes. 
This motivates us to seek the sparsest solution to $y = Ax$ by solving the following optimization problem:

\begin{eqnarray}
\hat{x}_0 = \arg \min ||x||_0 \mbox{    s. t.   }   Ax = y,
\end{eqnarray}
Here $||.||_0$ denotes the $\ell_0$-norm, which counts the number of nonzero entries in a vector. Solution to the above optimization problem provides the optimal solution, however, solving it for an underdetermined system is NP-hard. 

Recent development of the theory of compressive sensing and sparse representation has shown that if the solution $x_0$ is sparse enough, $\ell_1$-minimization provides same solution as that of $\ell_0$-minimization:
\begin{eqnarray}
\hat{x}_1 = \arg \min ||x||_1 \mbox{    s. t.   }   Ax = y,
\end{eqnarray}
In the past a large body of studies have successfully used $\ell_1$-minimization to find the sparse solution~\cite{rana2011adaptive,ear_phone_ipsn,east_ewsn,shen2013nonuniform,chou2012efficient,adaptivecompressive,shen2011non}. This problem can be solved in polynomial time by standard linear programming methods.

\subsection{Novel Extension of Sparse Random Classifier for Activity Classification}
\label{sec:dimensionalityReduction}
Recall from the previous section that $m<n$. Here, $n$ is the number of samples in an object instance. For large signals, for example for images, it is the number of pixels in the image vector after vectorization. For instance, if the face images are given at the typical resolution, 640$\times$480 pixels, the dimension $m$ is in the order of $10^5$. A higher dimension of the object offers extended processing complexity for regular computers, therefore, embedded platforms are out of question. In order to address this problem projections are taken to transform the images from image space to feature space:

\begin{eqnarray}
Ry = RAx
\label{eqn:dimensionReduction}
\end{eqnarray}
Here $R \in \mathbb{R}^{d\times n}$ is the so called projection matrix with $d<<n$.
In conventional SRC, Gaussian random numbers are used as the elements of $R$. However, we propose a novel construction of projection matrix. We first describe the projection matrix construction method in \cite{Carin:12}. The method assumes that the dictionary $A \in \mathbb{R}^{m \times n}$ and the number of projections $m$ are the inputs. The method first computes the singular value decomposition (SVD) of $A$:
\begin{equation}
D = U \Lambda V^T 
\label{sec:svdD}
\end{equation}  
where $^T$ denotes matrix transpose, $\Lambda \in \mathbb{R}^{n\times d}$ contains the singular values in its main diagonal, and $U \in \mathbb{R}^{n \times n}$ and $V \in \mathbb{R}^{d\times d}$ are orthonormal matrices. The method in \cite{Carin:12} is to \emph{randomly} choose $m$ columns from the matrix $U$. Let $\tilde{U}_m$ be a $n \times m$ matrix formed by these $m$ randomly chosen columns from $U$. The method is to use $ \tilde{U}_m^T$ as the projection matrix. The rationale of the method is that the columns in $U$ are highly uncorrelated with the dictionary $D$, therefore the sensing matrix $\Psi D = \tilde{U}_m^T D$ will have low coherence.

In our proposed method, we choose the $m$ columns in $U$ corresponding to the \emph{largest} $m$ singular values of $D$. We now explain why this is a better choice. To simplify notation, we assume that the SVD in \eqref{sec:svdD} has been permuted so that singular values appear in non-increasing order in the diagonal of $\Lambda$. With this notation, let $U_m$ denotes the sub-matrix containing the left-most $m$ columns of $U$; note that these $m$ columns correspond to the largest $m$ singular values of $D$. Our choice of projection matrix is therefore $U_m^T$. 

To understand why $U_m$ is a better choice, note that the activity classification problem can be stated as estimating the unknown coefficient vector $s$ from the projection $y$ by solving 
%\begin{align}
$y = R A x_0$.
%\end{align} 
We assume that the unknown coefficient vector $x_0$ comes from some probability distribution such that $\mathbb{E}[ x_0 x_0^T] = \mathbb{I}$ where $\mathbb{E}$ and $\mathbb{I}$ denote respectively the expectation operator and the identity matrix. 
It can be shown that the mean signal power $\mathbb{E}[y^T y]$ can be written as:
\begin{eqnarray}
\mathbb{E}[y^T y] = trace(R U \Lambda^2 U^T R^T)
\end{eqnarray} 
If we impose the constraint that each row of the projection matrix $R$ has unit norm, then the $R$ that maximizes $\mathbb{E}[y^T y]$ is given by the first $m$ rows of $U^T$ (or $U_m^T$), i.e. the $m$ left singular vectors corresponding to the largest $m$ singular values. This shows that our choice of projection matrix maximises the signal power of $y$. A higher signal power typically translates to lower estimation error.

\section{Occupancy Estimation Using Sensor Fusion}
\label{sec:sensorFusion}
When the phone is carried, it can be trivially determined that it is accompanied by its owner. However, in office people often tend to put the phone on desk and work.  Further, often they leave the desk while keeping in the phone on it. We consider determining occupancy in these two scenarios. We consider data from the \emph{phone accelerometer} and the \emph{phone microphone} to determine occupancy and use a number of assumptions to develop our occupancy estimation algorithm:
\begin{enumerate}
\item When people are typing it could create minor acceleration in the phone kept on the table.
\item When the person is not typing, the accelerometer may not be able to pickup the presence, however, sounds of general movement while seated can be picked up by the phone microphone.
\item When people are away from desk, the accelerometer can pickup acceleration from any devices, such as PC or laptop running on the desk. 
\end{enumerate}

\subsection{Features}
\subsubsection{Accelerometer} We study three axes of the accelerometer separately. We use a segment size containing one second of data. We investigate mean, 25 and 75 percentile and \emph{maximum} magnitude of each segment and do not observe any significant difference in classification for those choices. In the results section we use \emph{maximum} magnitude of each segment as a feature.  
%We plot the classification accuracy using accelerometer data in Fig.~\ref{}. 
\subsubsection{Sound} To preserve privacy we refrain using raw audio data for classification. We test the feasibility of signal energy and number of zero crossing as features and find number of zero crossing to be a better choice. Computation of signal energy involves determining a hamming window, which is a resource intensive operation and most importantly it does not provide classification accuracy as good as zero-crossing. We therefore use \emph{number of zero crossings} as a feature. We investigate on two attributes of audio data to calculate number of zero crossings: 1. \emph{sampling frequency} and 2. \emph{segment size}. We test a range of sampling frequencies spanning 8kHz,16kHz,32kHz and 48kHz and find that 48kHz provides the best results. We test window sizes of 5s, 15s, 25s, 35s and 45s and observe that 5s window provides the best results (Results are shown in Section~\ref{sec:results}).  

\subsection{Occupancy modeling}
Note that prior to applying fusion, we use the classical Support Vector Regression (SVR) for modeling occupancy individually from the audio and accelerometer features discussed in the previous section. The complete description of SVR is outside the scope of this paper. However, in this section we will provide intuition sufficient to understand the working principles of SVR. 
\subsubsection{Support Vector Regression} 
Our approach to predicting gait velocity is based on learning the functional relationship between the transition times and gait velocity.  To learn this relationship, we used a support vector regression model, which is widely used for prediction~\cite{rana2013feasibility,rana2011adaptive,rana2013passive}.

Consider a training set $\{(x_1,y_1),(x_2,y_2),...,(x_\ell,y_\ell)\}$, where $x_i$s are aggregated accelerometer or acoustic features and $y_i$s are the occupancy status. Support vector regression computes the function $f(x)$ that has the largest $\epsilon$ deviation from the actual observed $y_i$ for the complete training set.

\begin{figure}[ht]
\centering
\subfigure[No relation between $x$ and $y$.]
{
\includegraphics[width = 0.25\linewidth]{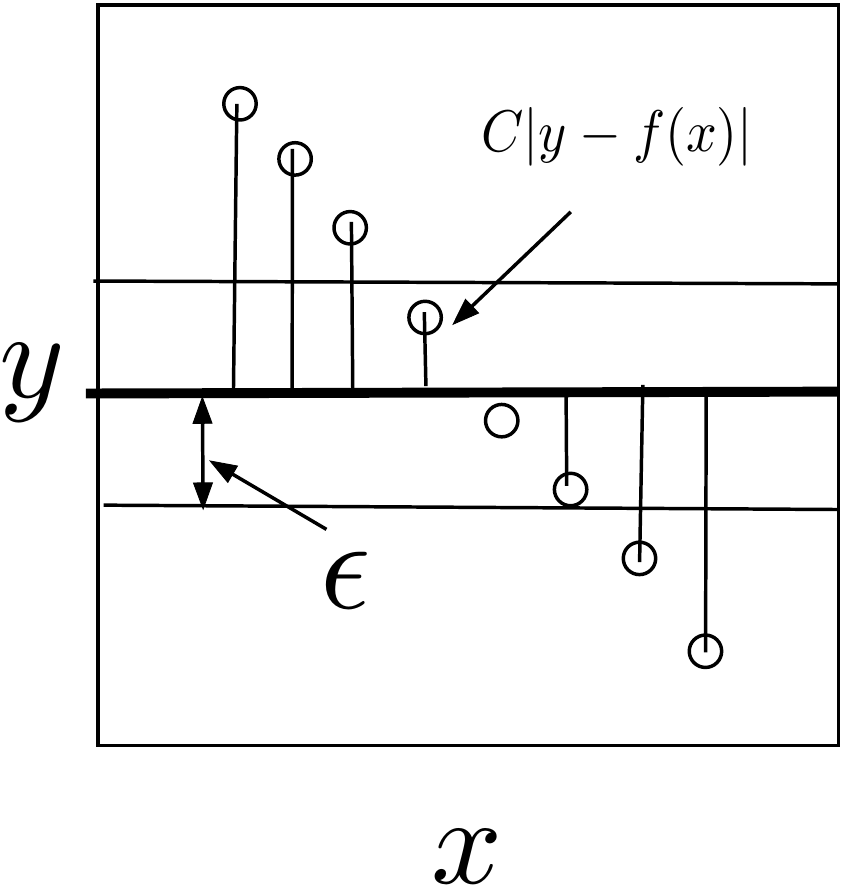}
\label{fig:svrFig2}
}
\subfigure[Linear regression.]{
\includegraphics[width = 0.25\linewidth]{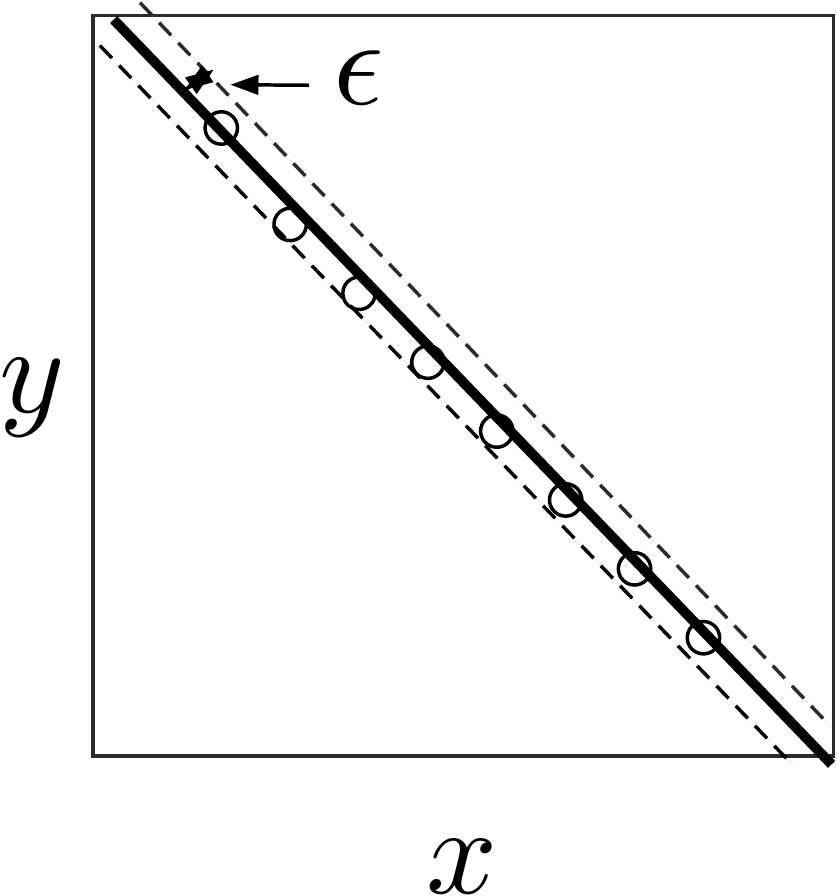}
\label{fig:svrFig1}
}
\subfigure[Linear SVR.]{
\includegraphics[width = 0.25\linewidth]{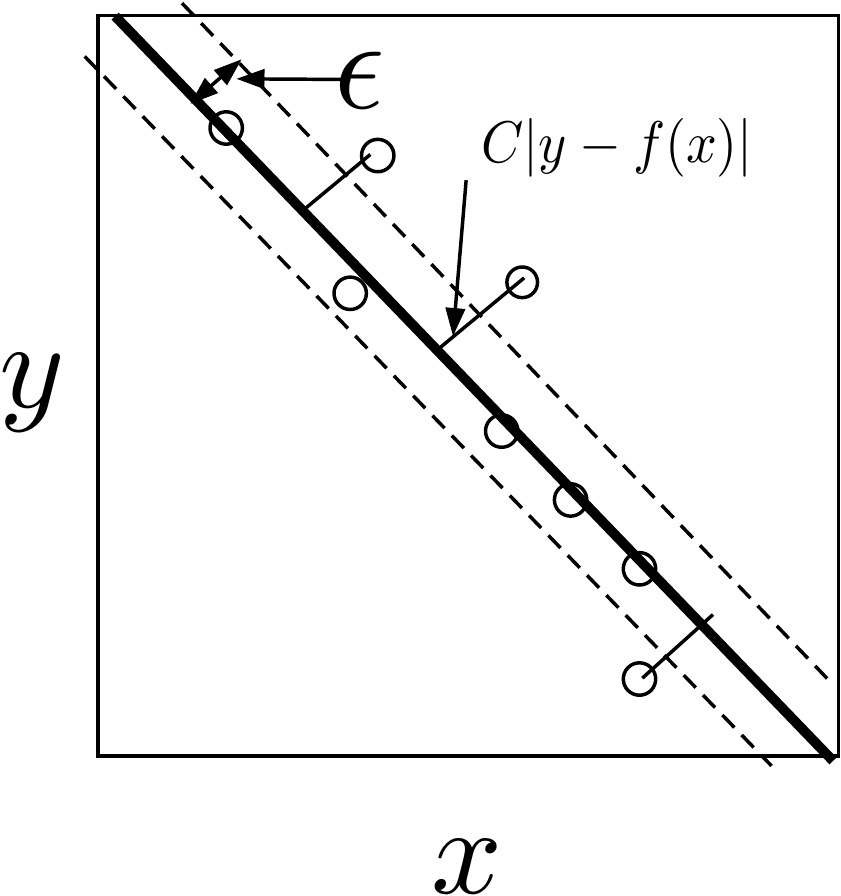}
\label{fig:svrFig3}
}
\caption{Support vector regression explained.}
\label{fig:svrExpalined}
\end{figure}

Let us assume the relationship between the variables is linear of the form $y = \omega x + b$, where $\omega$ (weight vector) and $b$ are parameters to be estimated. Fig.~\ref{fig:svrFig2} shows a few possible linear relationships between the points $x$ and $y$. The solid line in Fig.~\ref{fig:svrFig3} shows the SVR line given by $f(x) =  \omega x + b$. The cylindrical area between the dotted lines shows the region without regression error penalty. In the SVR literature this area is considered as the measure of complexity of the regression function used. Points lying outside the cylinder are penalized by an $\epsilon$-insensitive loss function~\eqref{eqn:lossFunction}~\cite{Vapnik:1995:NSL:211359} given by $|\xi|_\epsilon$.
\begin{eqnarray}
|\xi|_\epsilon := \begin{cases}
0 {\hspace{2 cm}		\mbox if } |\xi| \leq \epsilon\\
|\xi| - \epsilon			\mbox{ \hspace{1 cm}	 otherwise.}
\end{cases}
\label{eqn:lossFunction}
\end{eqnarray}
Now lets us explain the implication of a few different values of $\omega$. In the extreme case when $\omega = 0$ (as in Fig.~\ref{fig:svrFig2}), the functional relationship between $x$ and $y$ is least complex or in other words there is no relationship between $x$ and $y$. Therefore the overall error is very high. Next Fig.~\ref{fig:svrFig1} represents the case where the training data fits the solid line quite well. The solid line represents the classical regression analysis, where the loss function is measured as the squared estimation error. Note that although the solid line fits the data well, the cylindrical area between the dotted line is small, which means that the model will not generalize as well in predicting new data. SVR seeks to find a balance between the flatness of the area amongst the dotted lines and the number of training mistakes (see Fig.~\ref{fig:svrFig3}).

\begin{figure}
\centering
\includegraphics[width=0.7\linewidth]{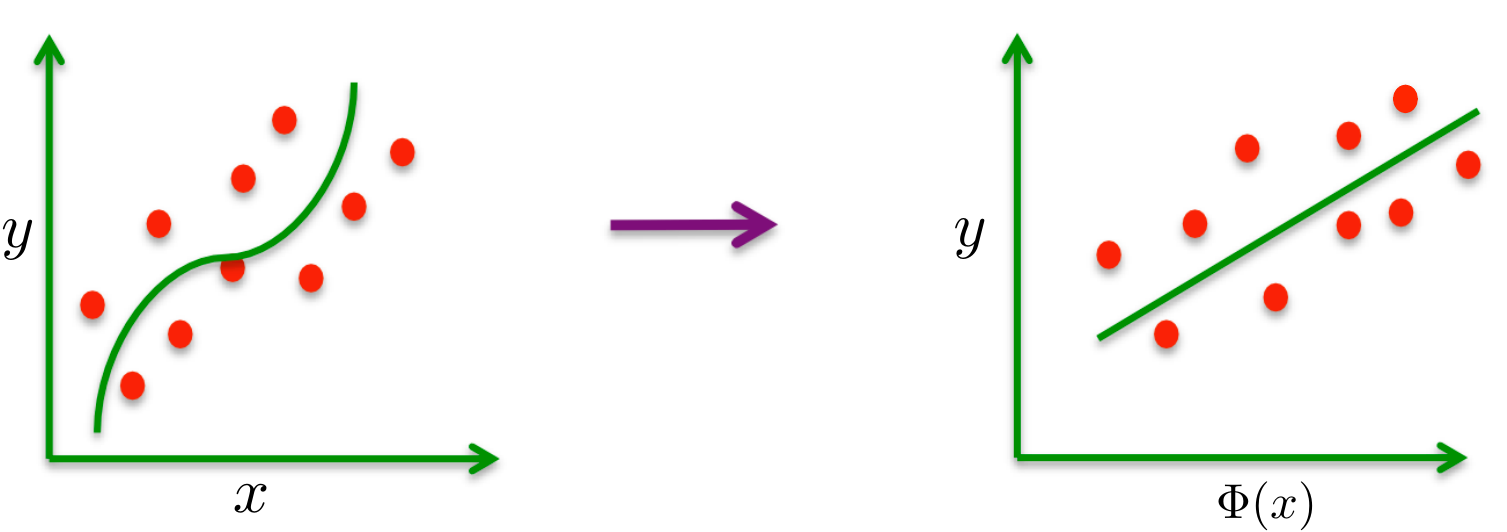}
\caption{Feature Space Transformation.}
\label{fig:transformSVR}
\end{figure}

Note that in many cases the relationship between the variables is non-linear as shown in the left diagram in Fig.~\ref{fig:transformSVR}. In those cases the SVR method needs to be extended, which is done by transforming $x_i$ into a feature space $\Phi(x_i)$. The feature space linearizes (right diagram in Fig.~\ref{fig:transformSVR}) the  relationship between $x_i$ and $y_i$, therefore, the linear approach can be used to find the regression solution. A mapping function or so called Kernel function is used to transform into feature space. There are four different functions which are frequently used as kernels within support vector regression: linear, RBF (Radial Basis Function), polynomial, and sigmoid. When the feature set is small, the RBF kernel is preferable over others. We use only one feature of transition time, therefore we use the RBF Kernel. However, we empirically verify that the RBF kernel performs better than the linear kernel. There are two parameters, namely $\gamma$ and $C$ (refer to~\cite{chang2011libsvm} for details) whose values need to be determined for best prediction. Here $C$ is the manually adjustable constant, and $\gamma$ is the kernel parameter which is formally defined as $K(x,y)= e^{-\gamma}||x-y||^2$. 
%The unknown parameters of the linear SVR $\omega$, $b$, and $\epsilon$ can be found as the unique solution of a dual of the primal problem  (see~\cite{smola1996regression}). There are a number of popular implementations of SVR in the literature including $\nu$-SVR~\cite{Schlkopf2000} and $\epsilon$-SVR~\cite{vapnik19196}. However, $\epsilon$-SVR or $\nu$-SVR just use different versions of the penalty parameter, the same optimization problem is solved in either case. We use $\epsilon$-SVR in our experiments. 
The overview of our SVR prediction framework is illustrated in Fig.~\ref{fig:transformSVR}.

%\begin{figure}
%\centering
%\includegraphics[width=1\linewidth]{finalFrameWork.pdf}
%\caption{Gait Velocity Prediction using SVR.}
%\label{fig:finalFrameWork}
%\end{figure}

\begin{figure}
\centering
\includegraphics[width=0.7\linewidth]{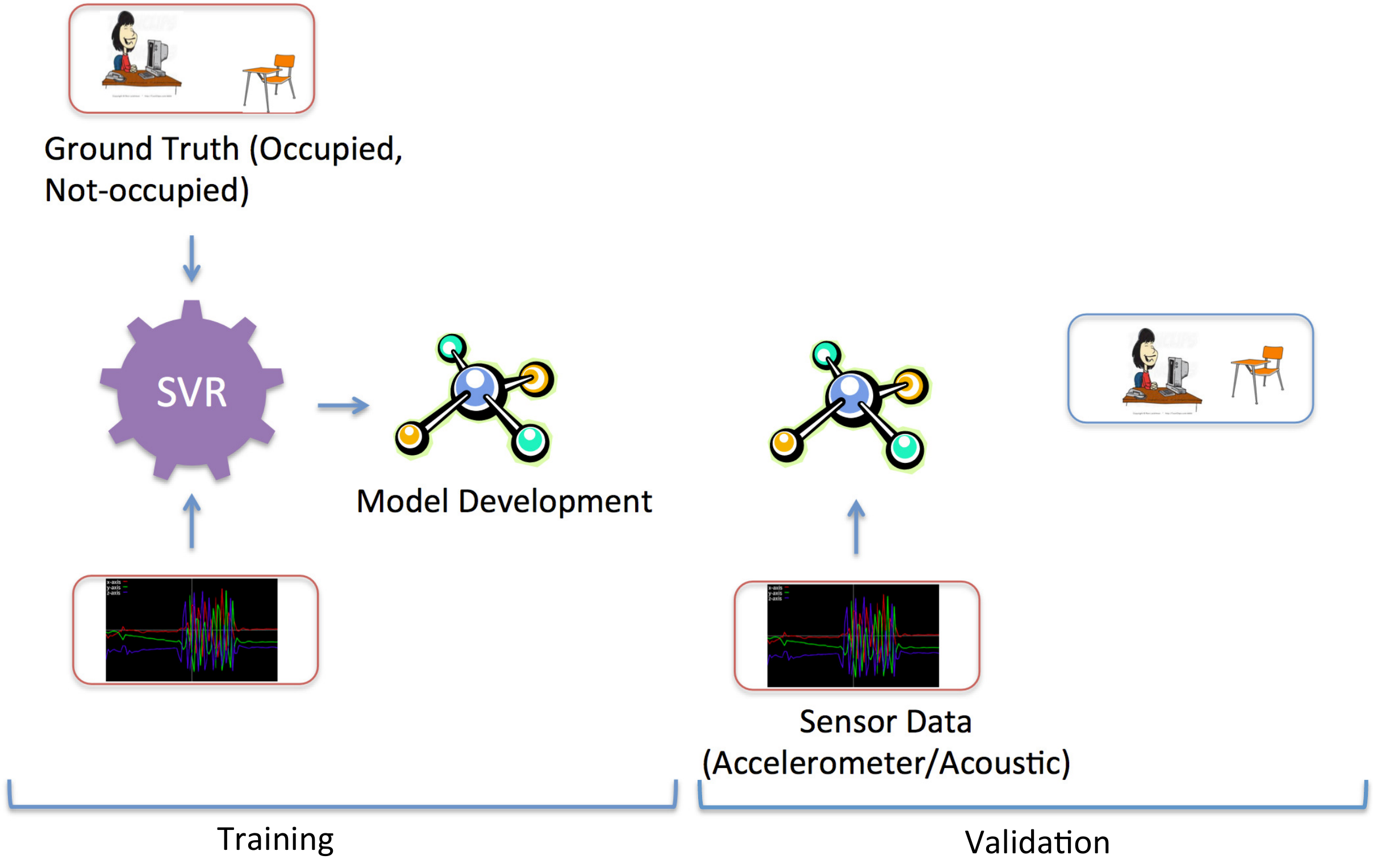}
\caption{SVR Prediction Framework.}
\label{fig:transformSVR}
\end{figure}
\subsection{Sensor Fusion} 
Since both of the sensor data are simultaneous collected on the phone, we investigate the feasibility of sensor fusion to improve the accuracy of occupancy estimation. Our fusion algorithm is motivated by the Dynamic Weighted Majority (DWM) Voting~\cite{kolter2003dynamic}, which is a ensemble learning algorithm. In DWM a series of learning algorithms, namely experts, are used to improve the predictive performance. At the beginning all the experts have equal weights, however, weights are penalized due to wrong prediction. We present a slight modification of the algorithm: we use multiple modalities as experts and use support vector regression for perdition, for each modality. Unlike DWM, we do not remove experts. Our extended Dynamic Weighted Majority Algorithm is presented in Algorithm~\ref{alg:dwm}. 

At the start of the algorithm both modalities have equal weight of 1. We use a 5-Fold cross validation where in each fold the accelerometer and acoustic features are used separately to predict the occupancy using the SVR model. If the prediction is wrong, weight is reduced by half ($\beta = 0.5$) otherwise weight is kept the same. At the end of these 5-Fold cross validation, weights for both modalities are learned and later on used for testing. We learn the weights  individually for each subject. 

%, shown in Figure 1, maintains as its concept description an ensemble of learning algorithms, each referred to as an expert and each with an associated weight. Given an instance, the performance element polls the experts, each returning a prediction for the instance. Using these predictions and expert weights, DWM returns as the global prediction the class label with the highest accumulated weight.

%The learning element, given a new training example, first

\begin{algorithm}
\caption{Extended Weighted Majority Algorithm}
\label{alg:dwm}
$\{x, y\}_1^n$: training data, feature vector and class label
\newline$\beta$ : factor for decreasing weights, $0 \leq \beta < 1$
\newline$c \in N^*$: number of classes
\newline $\{e, w\}_1^m$: set of sensing modalities and their weights
\newline $\sigma \in \mathbb{R}^c$: sum of weighted predictions for each class
\newline $\Lambda, \lambda \in {1, . . . , c}$: global and local predictions
\begin{algorithmic} 
\For{$i = 1,...,n$}
\State $\sigma_i \gets 0$
\For{$j = 1,.....,m$}
\State$\lambda \gets SVR(e_j,x_i)$
\If{($\lambda \neq y_i$)}
\State $w_j \gets  w_j \beta$
\EndIf
\State $\sigma_\lambda \gets \sigma_\lambda + w_j$
\EndFor
\State$\Lambda = \arg\max_j \sigma_j$
\EndFor
\end{algorithmic}
\end{algorithm}

\section{Results}
\label{sec:results}
\subsection{Activity Classification}
\subsubsection{DataSets}
We validate the performance of our proposed $\ell_1$ classifier, using publicly available dataset from University of California, Irvine (UCI)~\cite{anguita2012human}. In order to create this datasets, experiments were                                                                                                                                                                                                                                                                                                                                   carried out with a group of 30 volunteers within an age bracket of 19-48 years. Each person performed the six activities: \emph{standing, walking, laying, walking, walking upstairs} and \emph{walking downstairs}, wearing the  a Samsung Galaxy S2 smartphone on the waist. The features selected for this database come from the accelerometer and gyroscope 3-axial raw signals tAcc-XYZ and tGyro-XYZ. These time domain signals (prefix 't' to denote time) were captured at a constant rate of 50 Hz. Then they were filtered using a median filter and a 3rd order low pass Butterworth filter with a corner frequency of 20 Hz to remove noise. Similarly, the acceleration signal was then separated into body and gravity acceleration signals (tBodyAcc-XYZ and tGravityAcc-XYZ) using another low pass Butterworth filter with a corner frequency of 0.3 Hz. 

Subsequently, the body linear acceleration and angular velocity were derived in time to obtain Jerk signals (tBodyAccJerk-XYZ and tBodyGyroJerk-XYZ). Also the magnitude of these three-dimensional signals were calculated using the Euclidean norm (tBodyAccMag, tGravityAccMag, tBodyAccJerkMag, tBodyGyroMag, tBodyGyroJerkMag). 

Finally a Fast Fourier Transform (FFT) was applied to some of these signals producing fBodyAcc-XYZ, fBodyAccJerk-XYZ, fBodyGyro-XYZ, fBodyAccJerkMag, fBodyGyroMag, fBodyGyroJerkMag. (Note the 'f' to indicate frequency domain signals). In total there were \emph{561} features made available for this dataset.
The experiments have been video-recorded to facilitate the data labeling.

\begin{figure}
\centering
\includegraphics[width = 0.5\linewidth]{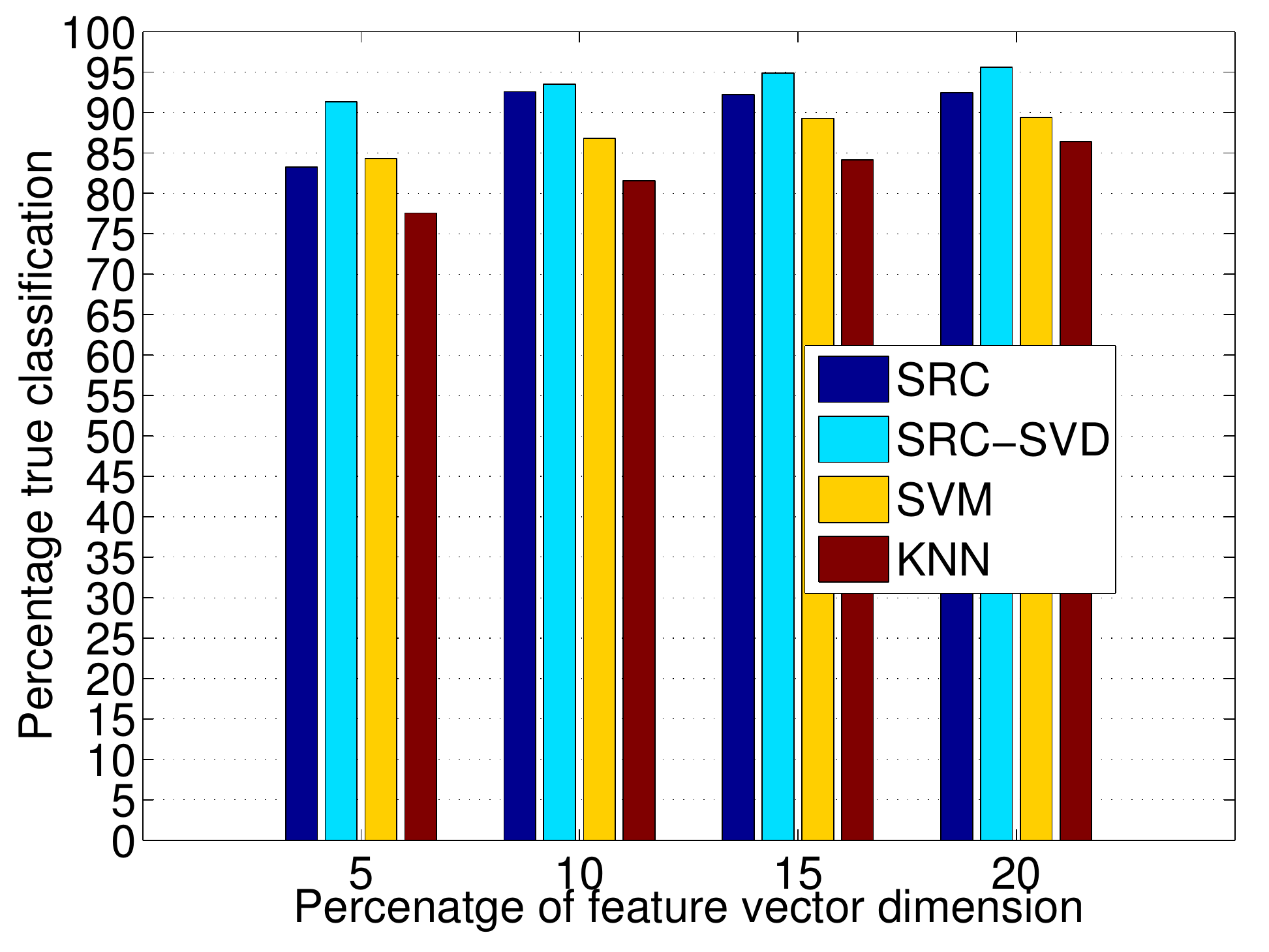}
\caption{Comparison of $\ell_1$-miimization approaches with SVM and kNN. The feature vector consists of 561 attributes. }
\label{fig:randomProjectionVersusSVM.eps}
\end{figure}

\begin{figure*}
\centering
\subfigure[]{
\includegraphics[width = 0.45\linewidth]{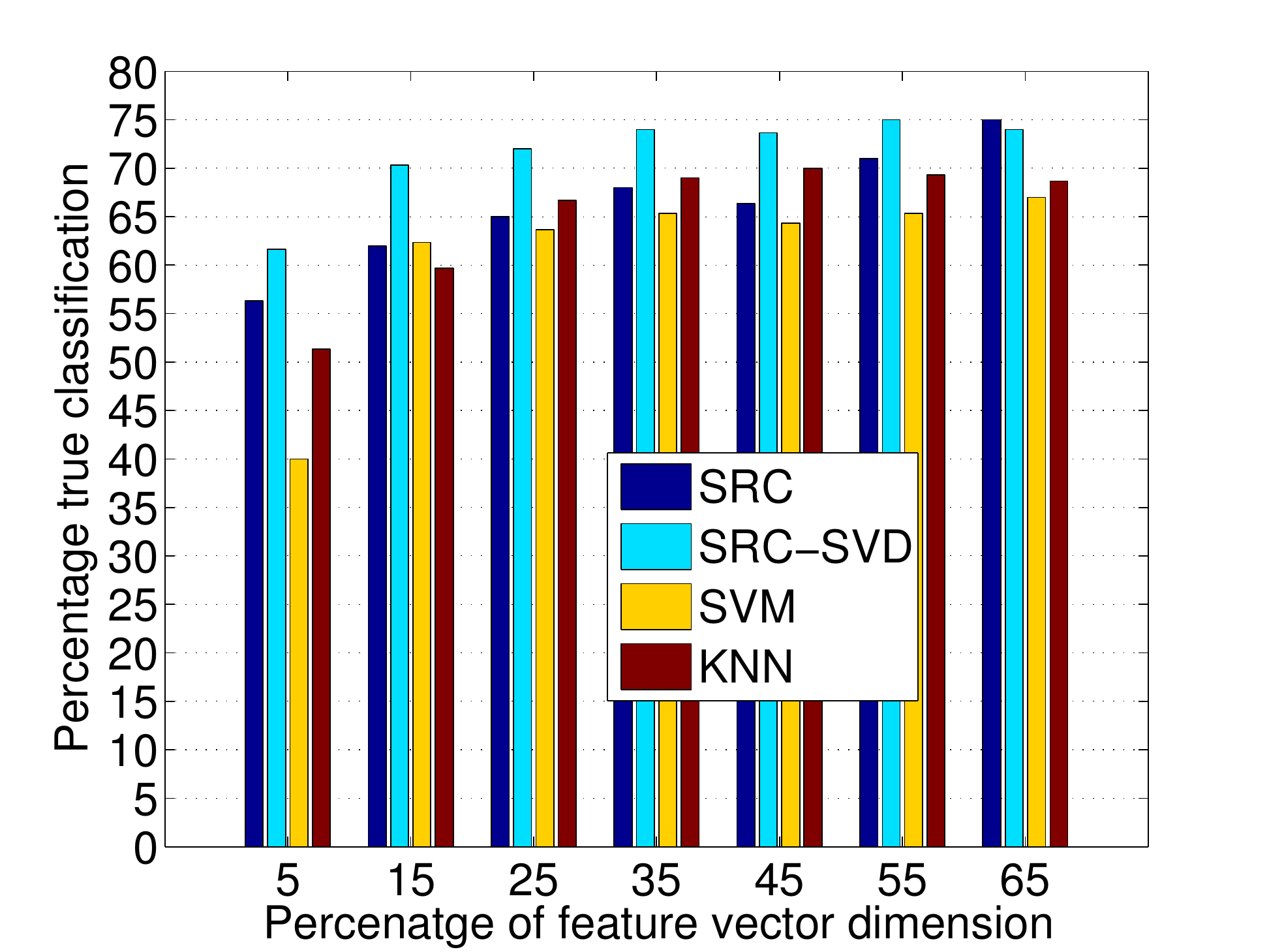}
}
\subfigure[]{
\includegraphics[width = 0.45\linewidth]{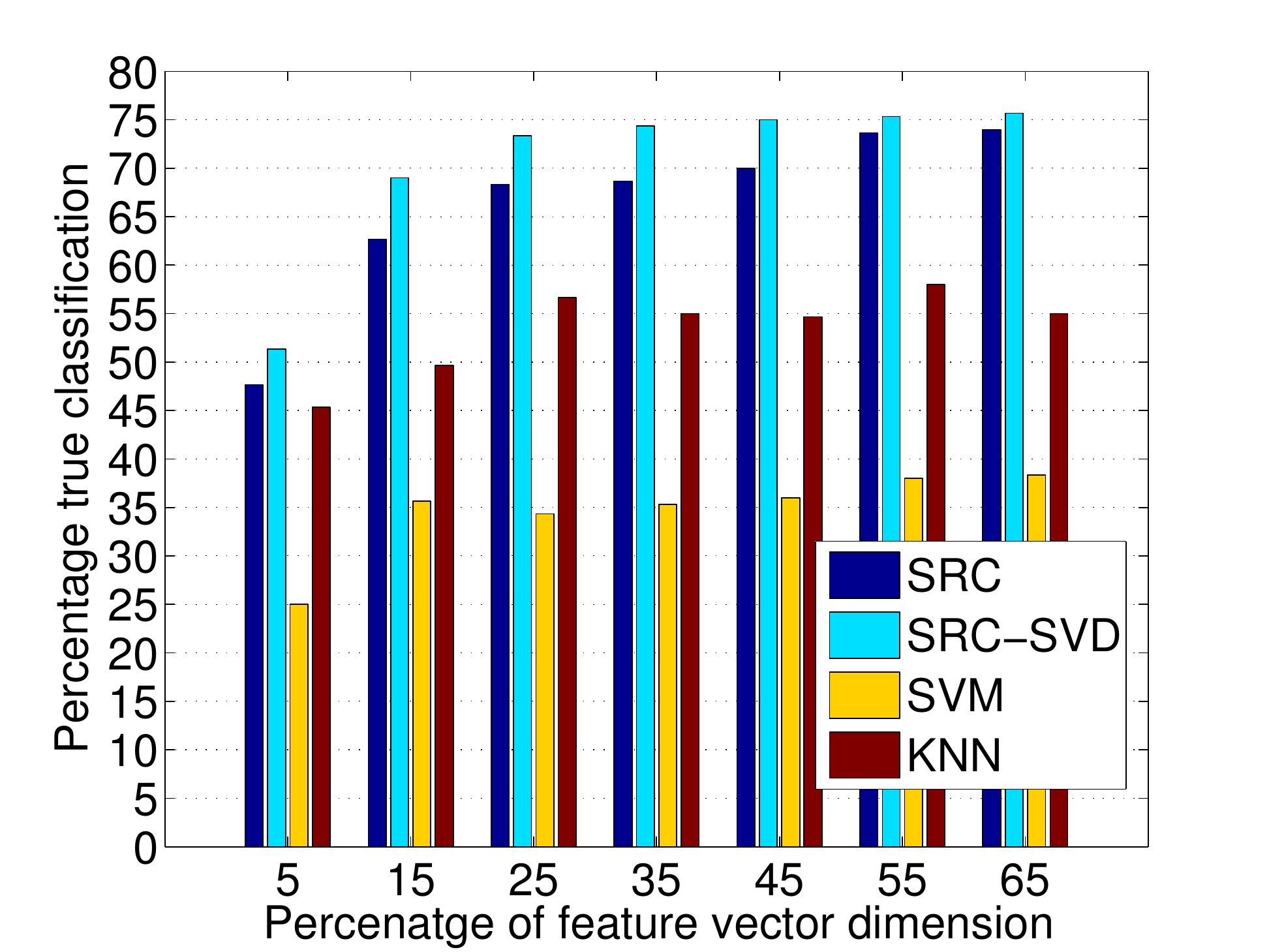}
}
\subfigure[]{
\includegraphics[width = 0.45\linewidth]{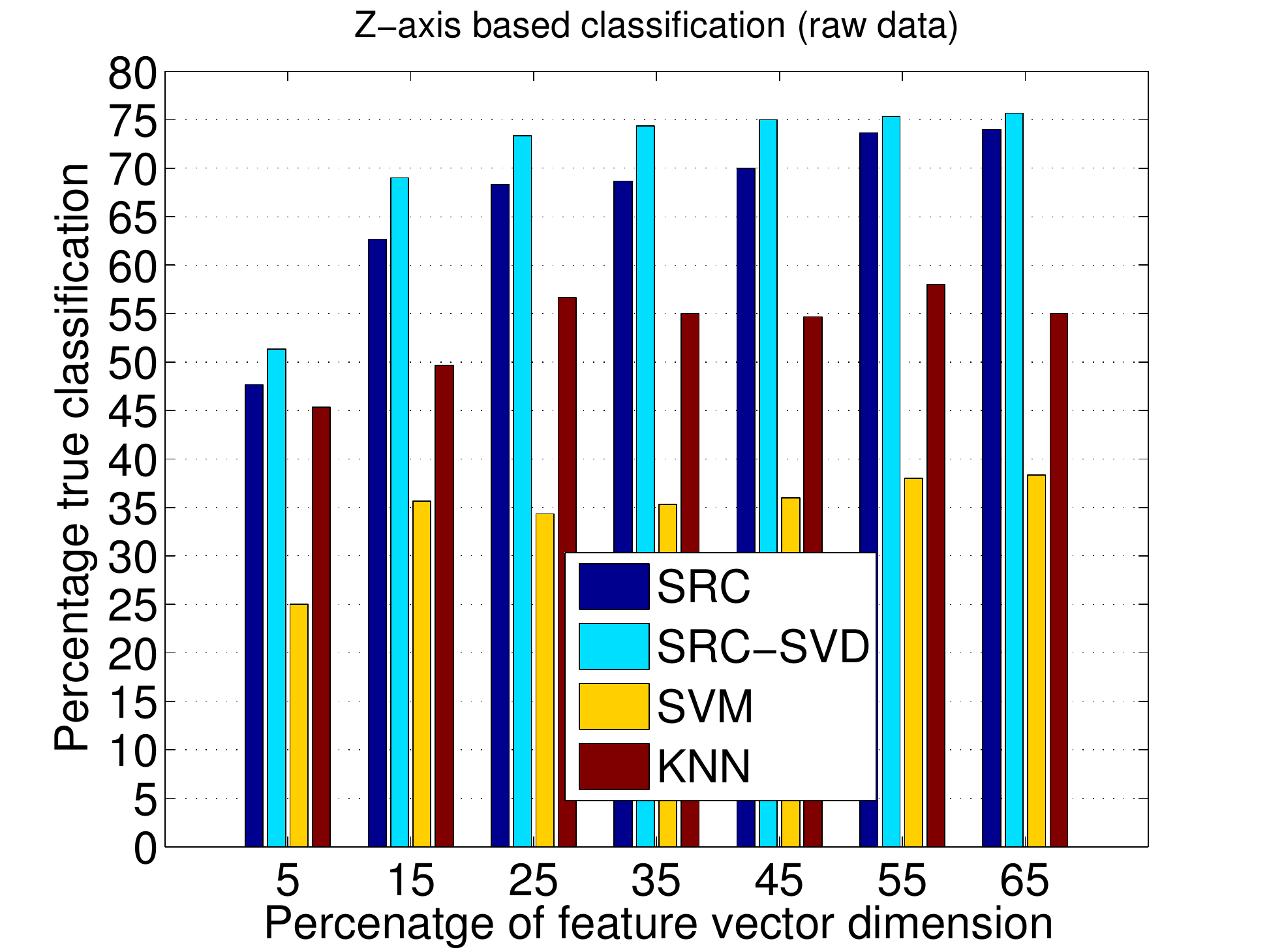}
}
\caption{Activity classification using raw data accelerometer data. Feature vector size is 128.}
\label{fig:rawDataClassification}
\end{figure*}
\subsection{ Activity Classification}
The key fundamental aspect of SRC is that it can perform classification with random features or in other words with raw sensor data. Therefore, it is possible to avoid the extensive feature extraction process. However, there is a trade-off between processing and accuracy while choosing between random and sophisticated features, which we present in Fig.~\ref{fig:randomProjectionVersusSVM.eps} and Fig.~\ref{fig:rawDataClassification}. We call our proposed extension of SRC, SRC-SVD. We benchmark the performance of SRC-SVD upon comparing with three other powerful alternatives: SRC (conventional implementation), Support Vector Machine (SVM) and k-nearest neighbor(kNN). 

In Fig~\ref{fig:randomProjectionVersusSVM.eps}, the 561 features are used for various classifiers. We apply dimension reduction (using \eqref{eqn:dimensionReduction}) to the signals. Due to our interest in smaller feature dimension, we present results for 95\% to 80\% of dimension reduction.  We have a number of observations from this figure: 1. in general our proposed SRC-SVD  performs the best compared to its alternatives. 2. SRC-SVD  performs the best compared to the alternatives when dimension reduction is 95\%. 3. Performance of SRC-SVD improves while dimension is increased from 5\% to 15\%, but plateaus beyond that point. The highest classification accuracy is 95\%.

In Fig.~\ref{fig:rawDataClassification}, we present the classification performance of SRC-SVD to classify using raw accelerometer data. We use all three axes of the accelerometer separately. Similar to Fig.~\ref{fig:randomProjectionVersusSVM.eps}, we also compare the performance of SRC-SVD with SRC, SVM and kNN. We observe that 1. SRC-SVD performs the best compared to the other classification methods. 2. X-axis provides the best classification commonly for all classification methods. 3. Classification performance of SRC-SVD increases gradually when dimension increases from 5\% to 35\% and plateaus beyond that point. The highest classification accuracy achieved is 75\%.
\subsection{ Occupancy Estimaiton}
\subsubsection{Experimental Settings}  
Experiments to determine occupancy were conducted with the assistance of three office colleagues. Each of them were given a Nexus 4S smartphone temporarily to collect data about their occupancy. They were allowed to use the phones for personal purpose (day to day usage). The subjects logged the information about their occupancy and location of the phone. We only used the log entries indicating phone on table and in/out of office to validate our occupancy algorithm. There was no information available about how the phone was placed on the table, such as how far from keyboard etc. 
\begin{figure*}
\centering
\subfigure[Occupied]{
\includegraphics[width = 0.45\linewidth]{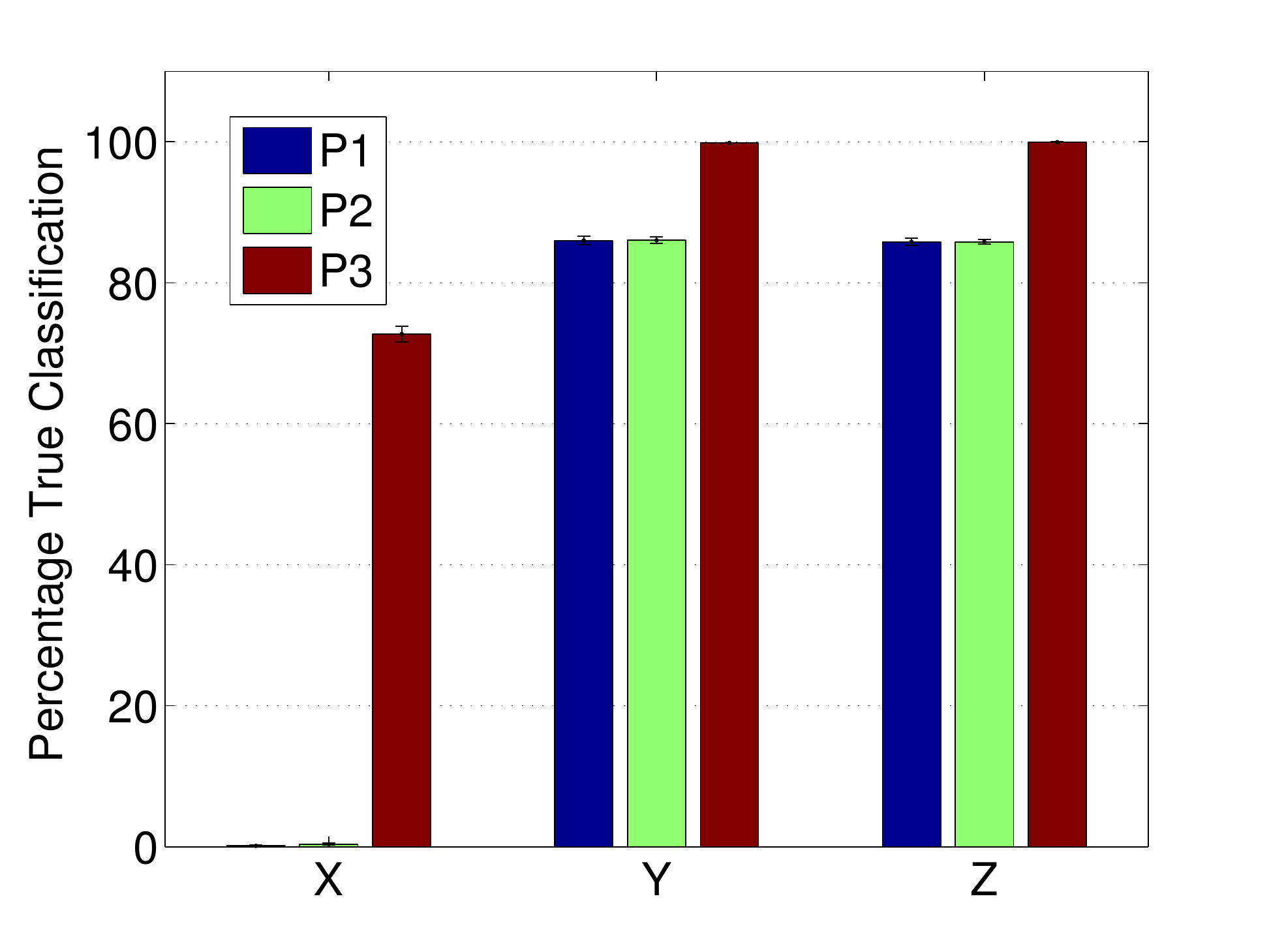}
\label{fig:presentWithAcc}
}
\subfigure[Unoccupied]{
\includegraphics[width = 0.45\linewidth]{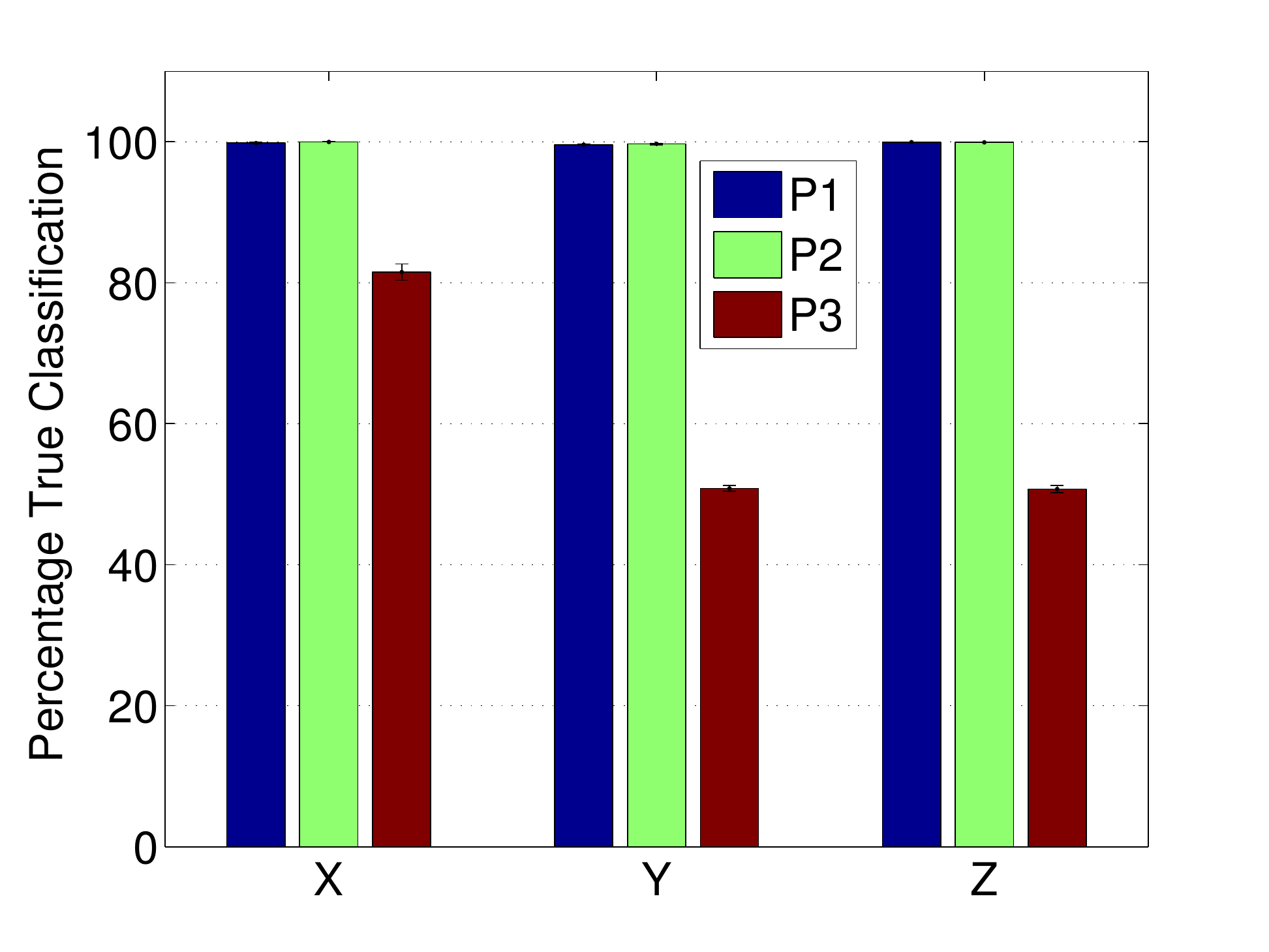}
\label{fig:absentWithAcc}
}
\caption{Occupancy estimation using {Accelerometer Data}. P1, P2 and P3 are three individual subjects. X, Y and Z corresponds to three axes of the accelerometer. The error bar shows the standard error of the mean.  \color{black}{The standard error of the mean now refers to the change in mean with different experiments conducted each time.}}
\label{fig:accuracyWithAccelerometer}
\end{figure*}

\begin{figure*}
\centering
\subfigure[Occupied]{
\includegraphics[width = 0.45\linewidth]{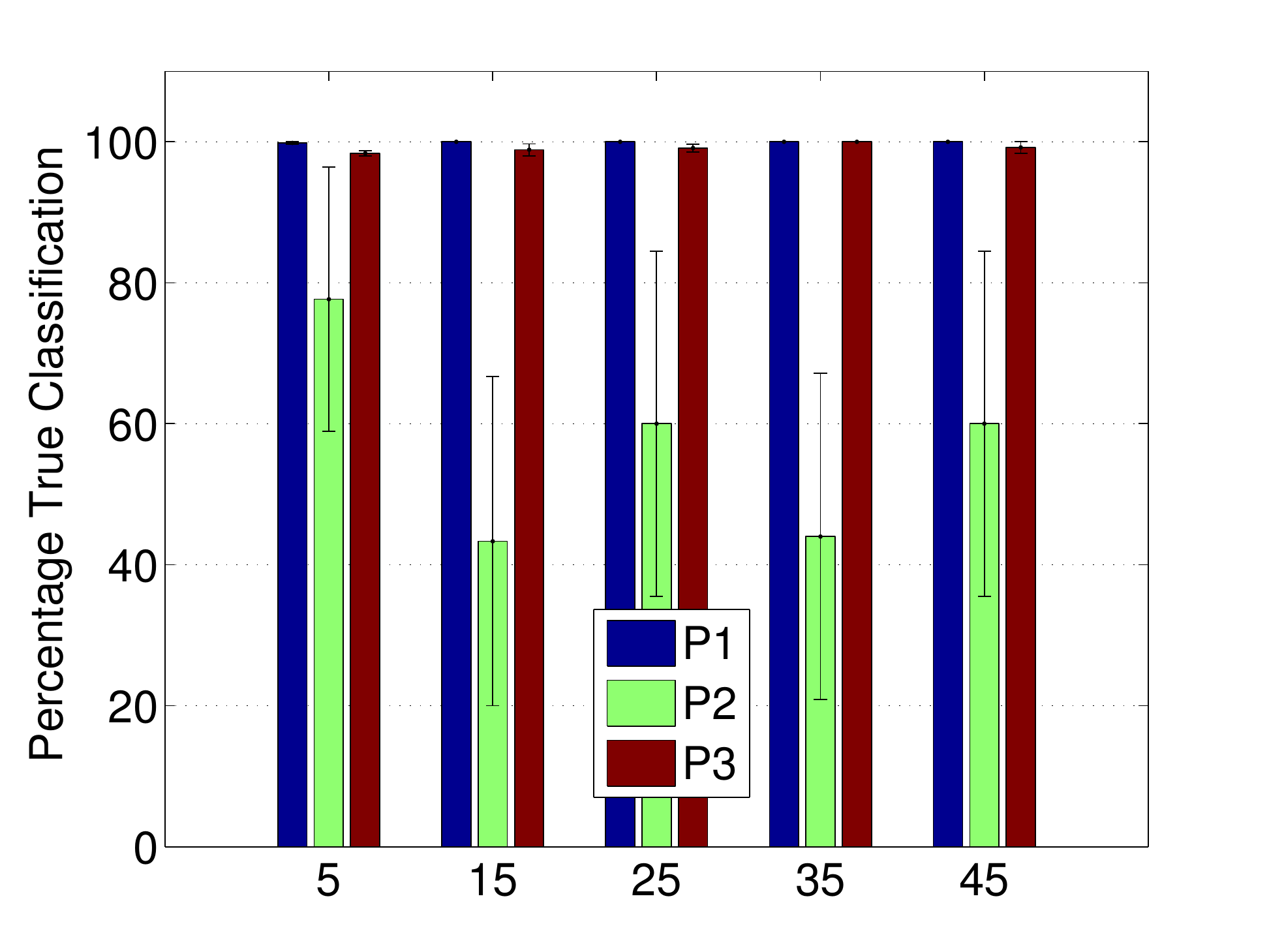}
\label{fig:presentWithSound}
}
\subfigure[Unoccupied]{
\includegraphics[width = 0.45\linewidth]{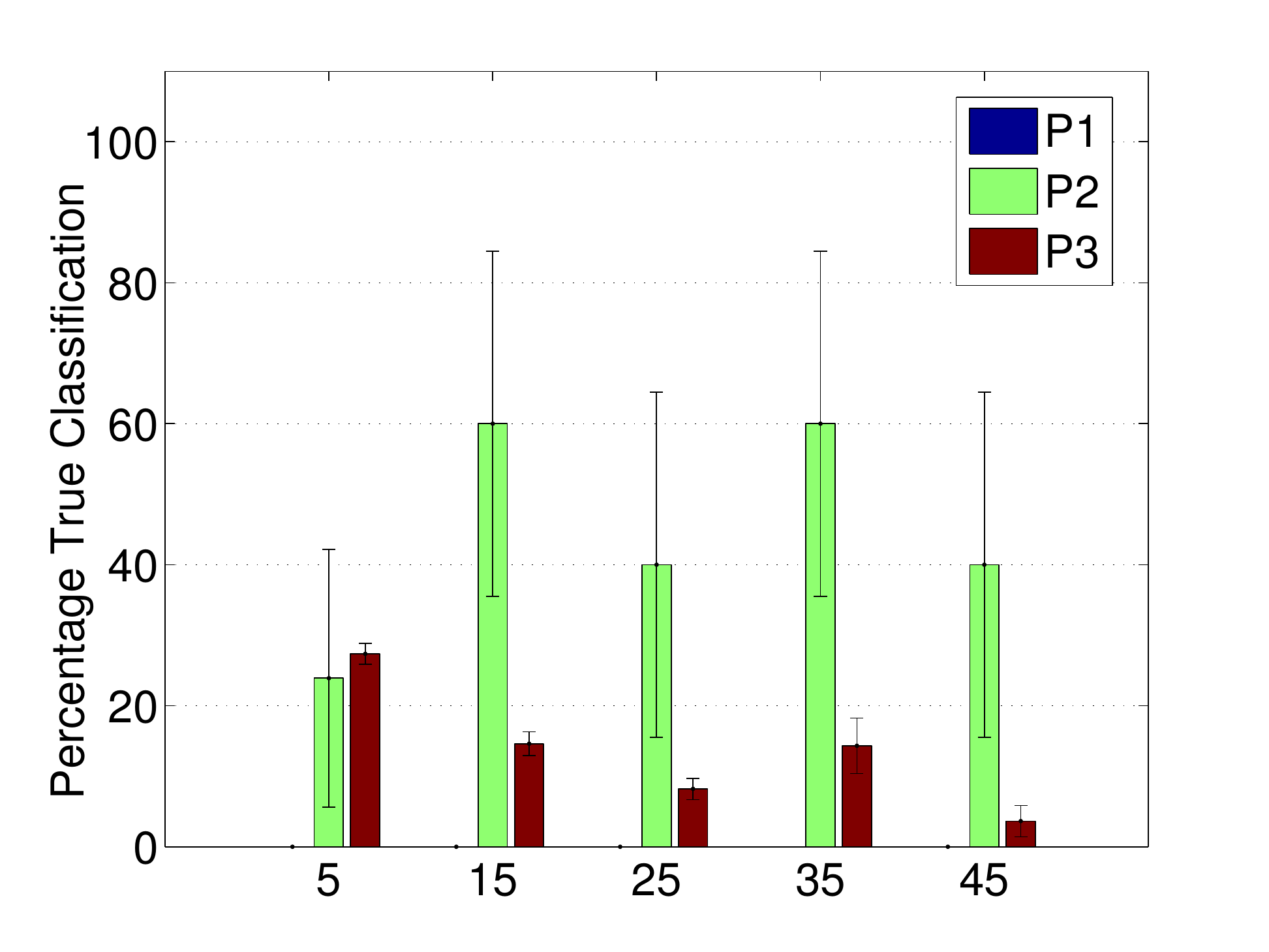}
\label{fig:absentWithSound}
}
\caption{Occupancy estimation using {Microphone Data}. P1, P2 and P3 are three individual subjects. X-axis shows the bin size. The error bar shows the standard error of the mean.} 
\label{fig:accuracyWithMic}
\end{figure*}

\begin{figure*}
\centering
\subfigure[Occupied]{
\includegraphics[width = 0.45\linewidth]{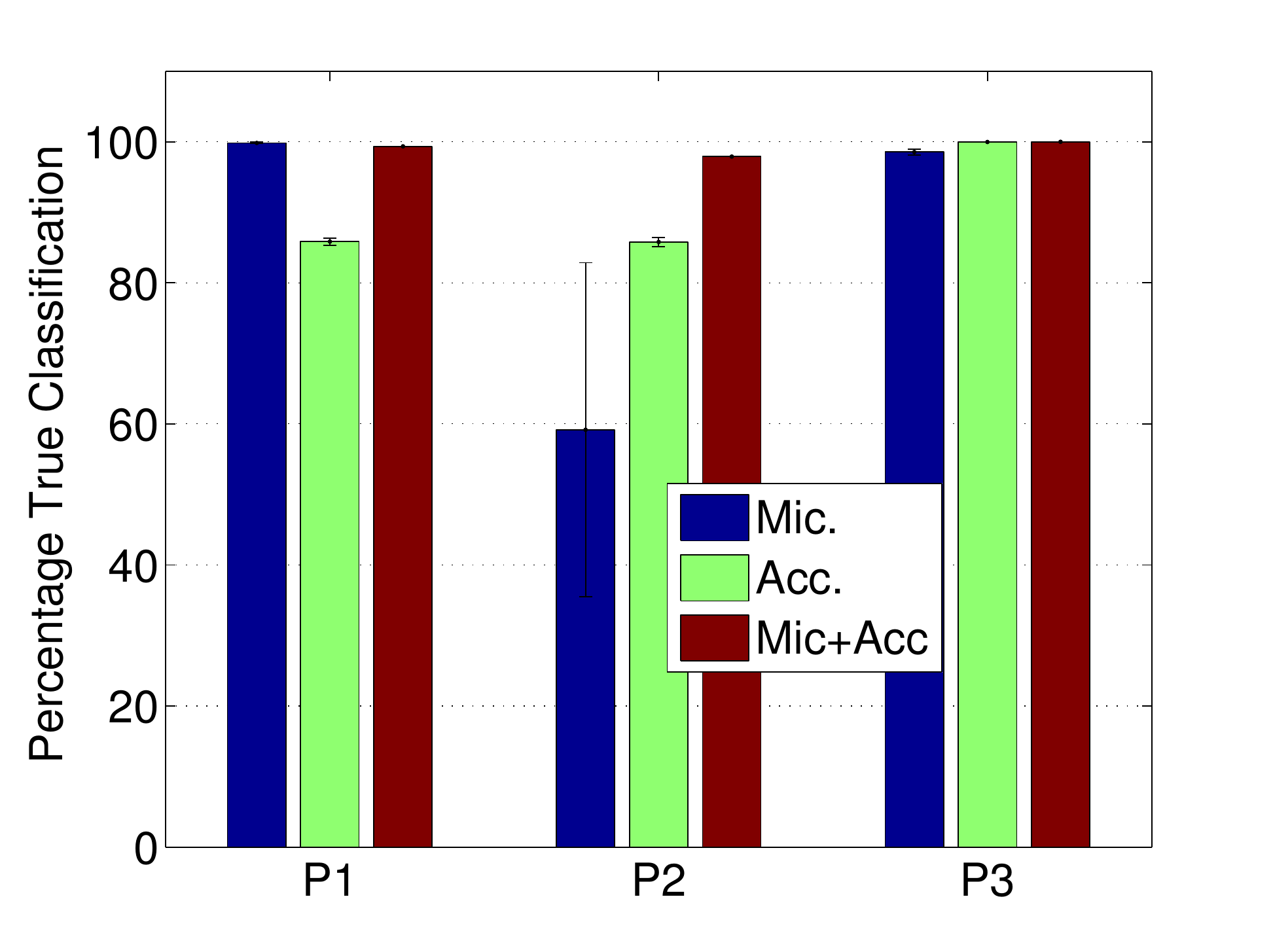}
\label{fig:presentWithSound}
}
\subfigure[Unoccupied]{
\includegraphics[width = 0.45\linewidth]{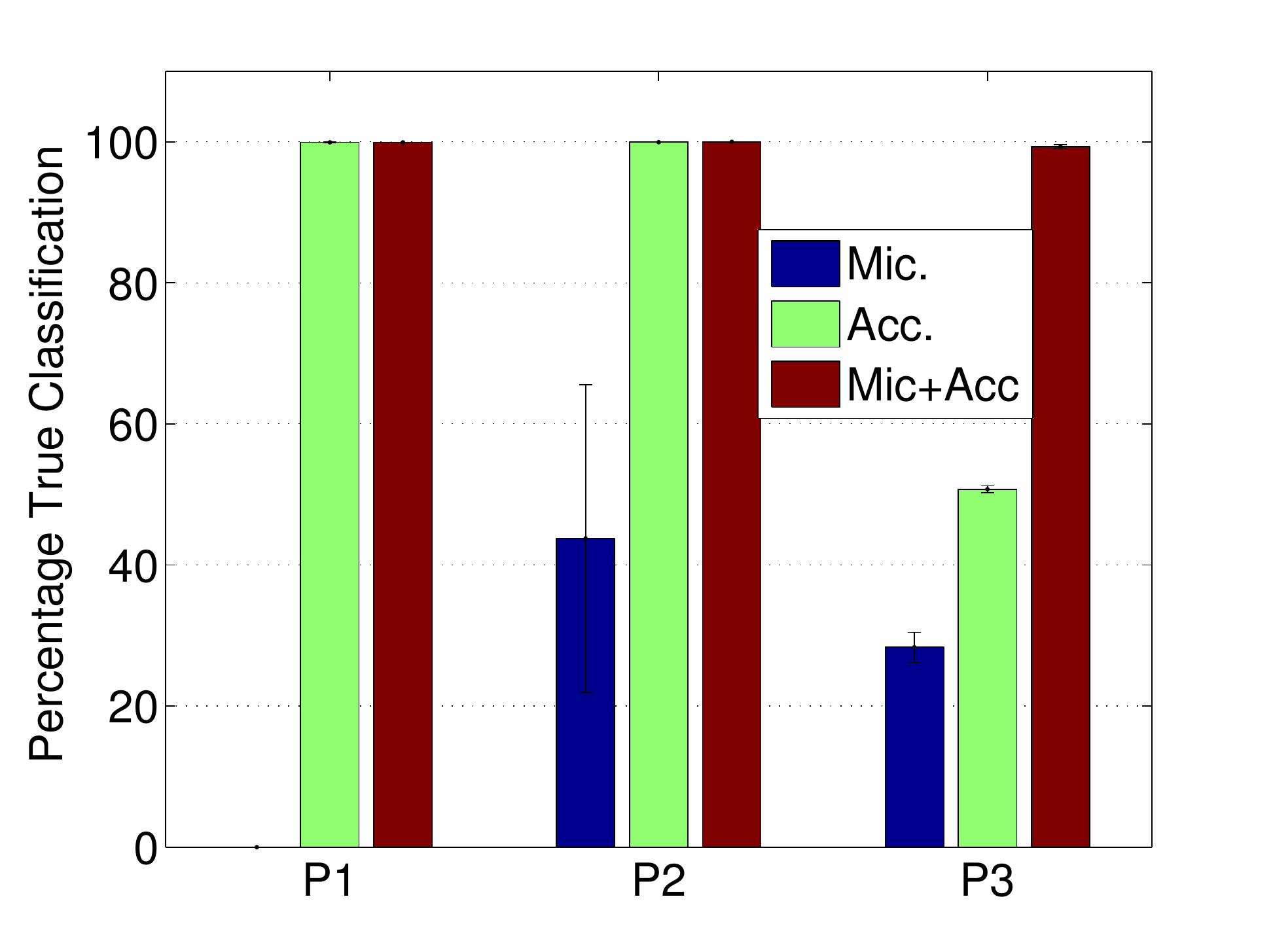}
\label{fig:absentWithSound}
}
\caption{Comparison amongst single and fused sensing modalities. Z-axis data of the accelerometer and 5s sampling window for microphone have been used. Weighted majority voting has been applied to fuse data from these two modalities. P1, P2 and P3 are three individual subjects. Acc. and Mic. are short form of accelerometer and microphone, respectively.} 
\label{fig:accuracyWithFusion}
\end{figure*}

\subsection{Occupancy Estimation Accuracy}
The results related to occupancy estimation accuracy are shown in Fig.~\ref{fig:accuracyWithAccelerometer} to Fig.~\ref{fig:accuracyWithFusion}. In Fig.~\ref{fig:accuracyWithAccelerometer} we present the accuracy when only accelerometer data is used. When the phone is on the desk, Z-axis is vertical to the desk surface. So, intuitively, most of the acceleration should be experienced along Z-axis. This is also reveled in this figure. For all our subjects, Z-axis provides better accuracy in both occupied and unoccupied cases.

Fig.~\ref{fig:accuracyWithMic} shows the occupancy estimation performance when using zero crossing as a feature of microphone data. The sound is recorded at 48KHz and we investigate the feasibility of various sampling window, such as 5s, 10s, 15s, and so on as shown in the figure. We observe that 1. microphone is particular suited to determine the occupied case. It performs badly for the unoccupied case and 2. the sampling window of 5s provides the best accuracy.

Given the differential performances of each modality, we investigate the feasibility of joint estimation using both modality. We define a fusion algorithm (Algorithm \ref{alg:dwm}) which is motivated by the Dynamic Weighted Majority Voting.  The sensor fusion provides significantly better performance compared to the individual modality in both occupied and unoccupied cases. It can be found from Fig.~\ref{fig:accuracyWithFusion} that for all three subjects and for both occupied and unoccupied cases, the accuracy is almost 100\%. We use z-axis data from accelerometer and use a 5s sampling window for the microphone data (sampled at 48KHz) in the sensor fusion algorithm.
%\begin{figure*}
%\centering
%\subfigure[]{
%\includegraphics[width = 0.65\linewidth]{summaryResultsAdvancedAccl}
%\label{fig:home.13}
%}
%\subfigure[]{
%\includegraphics[width = 0.45\linewidth]{summaryResultsAdvancedSoundPresent}
%\label{fig:ceilingImage.eps}
%}
%\subfigure[]{
%\includegraphics[width = 0.45\linewidth]{summaryResultsAdvancedSoundAbsent}
%\label{fig:walkingImage.eps}
%}
%\caption{Pi's are individuals. (a) Accelerometer estimated presence and absence. (b) 25 Percentile of Zero Crossing in audio data determining Presence(c) 25 Percentile of Zero Crossing in audio data determining Absence.}
%\label{fig:miscelleneous}
%\end{figure*}
%\

%We will show, through numerical evaluations in Section \ref{sec:eval}, that our method of computing projection matrix has two advantages.
%\begin{enumerate}
%\item For a given number of projections $m$, it reduces the trajectory reconstruction error.
%\item Because we use a deterministic way of computing the projection matrix, we get less variability in the error of trajectory reconstruction.
%\end{enumerate} 


\section{Conclusions}
\label{sec:conclude}
In this paper we have presented two novel methods for activity classification and occupancy estimation, which are two crucial components for intelligent HVAC control. Physical activities performed recently typically indicate the need for adjusting thermal preferences. Similarly, occupancy is very important in HVAC control. It can save power consumption substantially while maintaing thermal comfort if the HVAC duty cycling can be adapted correctly to the occupancy status. 

The novelty of our activity classification algorithm lies in the extension of the Sparse Random Classifier. We introduce a novel construction of projection matrix that offers higher accuracy compared to the conventional implementation of Sparse Random Classifier (SRC). We report that the classification accuracy can be as high as 95\% while using 50\% smaller feature dimension compared to the existing implementation of SRC.

Our occupancy estimation is novel because, for the first time we propose the fusion of accelerometer and microphone data from smartphone to determine occupancy. We perform experiments with real subjects and experimental results reveal that we achieve almost 100\% classification accuracy, which is substantially high. In these experiments accelerometer and microphone data were used when the subject reported that phone was on desk. One of the most important aspects of our experiments is that it was completely uncontrolled. There was no information available on how the phone was placed on the desk, i.e., how far it was placed from the keyboard or hand etc.

In our future studies, we want to validate the proposed sensor fusion algorithm with large number of subjects. In addition, we are also aiming to implement the activity classification algorithm on smartphone and conduct experiments with human subjects to validate its performance. 

\section{References}
\bibliographystyle{elsarticle-num} 
\bibliography{occuActivity.bib}



%\bibitem{}
%
%\end{thebibliography}
\end{document}